\documentclass[[journal,11pt,onecolumn,final]{IEEEtran}
\IEEEoverridecommandlockouts
\usepackage{url}
\usepackage{subfig}
\usepackage{cite}
\usepackage{amsmath,amssymb,amsfonts}
\usepackage{graphicx}
\usepackage{textcomp}
\usepackage{adjustbox}
\usepackage{xcolor}
\usepackage[textsize=tiny,textwidth=1.4cm]{todonotes}
\usepackage{mathtools,lipsum, nccmath,cuted}
\usepackage{siunitx}

\usepackage{comment}
\usepackage{makecell}
\usepackage{algpseudocode}
\usepackage{gensymb}
\usepackage{ulem}
\usepackage{colortbl}
\usepackage{hyperref}
\usepackage{enumitem}
\usepackage{booktabs} 
\definecolor{Gray}{gray}{0.925}
\usepackage[textsize=tiny]{todonotes}

\graphicspath{ {figures/} }
\def\BibTeX{{\rm B\kern-.05em{\sc i\kern-.025em b}\kern-.08em
    T\kern-.1667em\lower.7ex\hbox{E}\kern-.125emX}}
    
\usepackage{pgfplots}
\usepgfplotslibrary{colorbrewer}
\pgfplotsset{compat = 1.14, cycle list/Set1-8} 
\usetikzlibrary{pgfplots.statistics, pgfplots.colorbrewer,pgfplots.dateplot,
        shadows,
        matrix} 

\usepackage{pgfplotstable}
\usepackage{filecontents}
\usepackage{graphicx}
\pgfplotsset{compat=1.14}

\definecolor{blueLine}{RGB}{57,106,177}
\definecolor{blueFill}{RGB}{114,147,203}
\definecolor{redLine}{RGB}{204,37,41}
\definecolor{greenline}{RGB}{0,250,0}
\definecolor{blackLine}{RGB}{0,0,0}
\definecolor{goldLine}{RGB}{160,82,45}

\usepackage[font=footnotesize]{caption}
\begin{document}
\title{Automatic AI controller that can drive with confidence: steering vehicle with uncertainty knowledge}
\author{
    \IEEEauthorblockN{ Neha Kumari\IEEEauthorrefmark{3},Sumit Kumar\IEEEauthorrefmark{4},Sneha Priya\IEEEauthorrefmark{1},  Ayush Kumar\IEEEauthorrefmark{2}, Akash Fogla \IEEEauthorrefmark{4}\\}
    \IEEEauthorblockA{\IEEEauthorrefmark{1}ICFAI University, \IEEEauthorrefmark{2} Amity University, \IEEEauthorrefmark{3} University of Georgia,\IEEEauthorrefmark{4}Georgia State University}
}


\maketitle

\section{\textbf{\textit{ABSTRACT}}}
\textit{ In safety-critical systems that interface with the real world, the role of uncertainty in decision-making is pivotal, particularly in the context of machine learning models. For the secure functioning of Cyber-Physical Systems (CPS), it is imperative to manage such uncertainty adeptly. In this research, we focus on the development of a vehicle's lateral control system using a machine learning framework. Specifically, we employ a Bayesian Neural Network (BNN), a probabilistic learning model, to address uncertainty quantification. This capability allows us to gauge the level of confidence or uncertainty in the model's predictions.
The BNN based controller is trained using simulated data gathered from the vehicle traversing a single track and subsequently tested on various other tracks. We want to share two significant results: firstly, the trained model demonstrates the ability to adapt and effectively control the vehicle on multiple similar tracks. Secondly, the quantification of prediction confidence integrated into the controller serves as an early-warning system, signaling when the algorithm lacks confidence in its predictions and is therefore susceptible to failure.
By establishing a confidence threshold, we can trigger manual intervention, ensuring that control is relinquished from the algorithm when it operates outside of safe parameters.
}
\subsection*{Keywords}
\textbf{\textit{Bayesian neural networks, Probabilistic modeling,  Torcs, surrogate modeling, Electronics Control Unit}}


\section{Introduction}
\label{sec:introduction}

Self-driving vehicle comprises intricate machine learning (ML) and artificial intelligence (AI) components, functioning as subsystems within a sophisticated driving system, including a perception system and obstacle detection system. Although prevalent self-driving systems predominantly employ multiple ML and AI models for specialized tasks, the paradigm of end-to-end driving systems is relatively scarce \cite{bojarski2016end}.
ML and AI have demonstrated significant efficacy in control systems\cite{abbeel2010autonomous,vardhan2021rare}, predictive modeling for complex phenomena such as cancer detection and stock market trends\cite{al2019comparative,ghazanfar2017using}, and automated design processes\cite{vardhan2022deepal}. This study specifically focuses on the utilization of two widely adopted ML models to govern lateral vehicle motion, with the overarching goal of scrutinizing their performance under conditions of abundant data availability. The ensuing comparison aims to serve as a foundational reference, furnishing insights for AI-based control designers regarding the judicious selection of models and the discernment of respective strengths and weaknesses.

For the lateral control of the vehicle, the TORCS open-source car racing simulator\cite{wymann2000torcs} is employed. The data set for this investigation is generated across diverse speed conditions through the utilization of a conventional Proportional-Integral-Derivative (PID) controller, with an emphasis on steering control during the training phase. The dataset encompasses distance measurements procured from a Light Detection and Ranging (LIDAR) sensor suite, concomitant with corresponding steering values.
A deep learning models, namely Bayesian Neural network, is selected for training and subsequent testing under different scenarios. These are the following contributions:
\begin{enumerate}
    \item The Bayesian Neural network-based controller manifests superior generalization capabilities in instances where access to an extensive dataset is constrained. It also provide the uncertainty in the prediction. 
    \item The uncertainty is used to decide when the model is not confident, and the steering control is overtaken by the manual driver. 
    \item When deployed on other similar tracks, the BNN controller successfully drives on the track without incident. When deployed on the more complex and convoluted track with hairpin curves and very twisted turns the uncertainty estimation helped to take manual control when uncertainty was beyond beyond threshold. It helped in safe navigation even on unseen terrains. 
\end{enumerate}

The rest of the paper is organised as follow. Section \ref{sec:methods} formulate the problem and provide background, approach and training details of controller. Section \ref{sec:experimentalResults} provides the experimental results of the experimentation. The related work is discussed in section \ref{sec:relatedWorks} and at the end we produce our conclusion and future direction of research\ref{sec:conclusionFutureWork}.

\section{Problem Formulation and Approach}
\label{sec:methods}
\subsection{Problem Formulation}
To formalize the problem, we introduce notations. Let $x$ denote the state of the vehicle estimated by sensors, sampled from an unknown distribution $P_x$. The objective of controller design is to learn a nonlinear function $f$ mapping $x$ to the control action $c$:

$$f: x \mapsto c \; \text{where} \; x \sim P_x$$

We consider the widely used architecture i.e. deep neural networks based probabilistic network called Bayesian Neural network as an instances of the architecture class $\mathcal{A}$. The choice of $f$ is due to two constraints: first- we want to learn a highly nonlinear control action and second we wnat some uncertainty in prediction.   We treat this problem as a supervised learning task, utilizing a labeled dataset $D = \{<x_i, c_i>\}$ for training. The dataset is split into training ($D_{train}$) and test ($D_{test}$) sets. The learning goal is to identify parameters distribution (for neural network) that effectively represent the training data and perform well on test data and can also measure the uncertainty in the prediction.

\subsection{Approach}
\begin{figure}[tbhp]
\centering
   \includegraphics[width=0.8
   \linewidth]{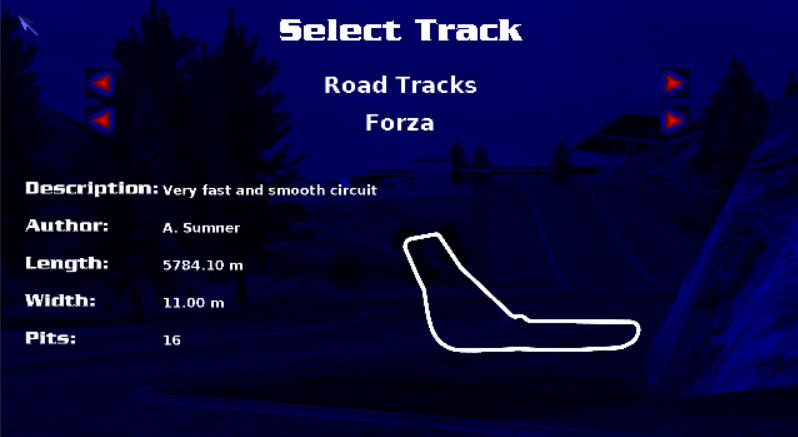}
   \caption{track: Forza}
   \label{fig:forza}
\end{figure}

In this section, we prove the background literature on the components used in the experiment. 
\textbf{Torcs Simulator} TORCS (The Open Racing Car Simulator) is an open source 3D racing car simulator that can be use to select and drive a car on selected track and visualise and log data using a peripheral  device or via a design controller (AI-based or traditional methods). All major aspects of vehicle dynamics can be accurately simulated, including mass, rotational inertia, impacts, suspension dynamics, linkages, differentials, friction and aerodynamics. Simplifying, it perform physical simulations using Euler integration of dynamical equations using a time discretization level of $0.002$ seconds. TORCS offers a variety of tracks and cars, sensor and road types. It is widely used in research community for simulation.
\\
\textbf{Data generation} For data generation purposes, \cite{fogla2023vehicle} used the TORCS' asset car and drove it on the 'Forza' road track (fig. \ref{fig:forza}) in order to generate data. For sensing the environment, Fogla et al \cite{fogla2023vehicle} deployed multiple LIDAR sensors attached to the car that can sense the track and the wall. For data generation purposes, a a tuned PID controller is designed that can drive the car successfully on a given speed on the track Forza. The data set consists of input as a set of distances measured by various LIDAR sensors and output as the steering value. By the finely tuned PID controller, multiple rounds of run is conducted on this track and the corresponding LIDAR distance data and the steering values are collected. Our model is trained on this labeled data set. 
 In our case, we want to train a regression model that estimate and predict the steering value on the LIDAR distance data during test and run time.  
\\

\textbf{Bayesian neural network, parameters and training }
Bayesian Neural Networks (BNNs) combine the principles of Bayesian statistics and inference with neural networks. Bayesian inference is a method of statistical inference in which Bayes' theorem is used to update the probability for a hypothesis as more evidence or information becomes available. Bayes' theorem is stated mathematically as:

\[ P(H|E) = \frac{P(E|H) \cdot P(H)}{P(E)} \]

where:
\begin{enumerate}
    \item  \( P(H|E) \) is the probability of hypothesis \( H \) given the evidence \( E \), known as the posterior probability.
    \item   \( P(H) \) is the probability of hypothesis \( H \) being true, known as the prior probability.
    \item  \( P(E|H) \) is the probability of observing evidence \( E \) given that hypothesis \( H \) is true, known as the likelihood.
    \item  \( P(E) \) is the probability of observing evidence \( E \), known as the evidence or marginal likelihood.
\end{enumerate}

A Bayesian Neural Network (BNN) extends this approach by placing a prior probability distribution over the weights of the network, rather than assigning fixed values in case of classical feed forward deep neural network. This means each weight is considered as a random variable described by a probability distribution, reflecting our uncertainty about its true value.
In a BNN, the forward pass involves sampling weights from their respective distributions and using these weights to make predictions. For a given input \( x \), the output \( y \) is computed as follows:
\[ y = f(\mathbf{x}, \mathbf{W}) \]
where \( f \) is the neural network function and \( \mathbf{W} \) represents the set of weights and biases sampled from their distributions. 
Learning in BNNs is about updating the distributions of the weights. This is done through Bayes' theorem by calculating the posterior distribution of the weights given the data:

\[ P(\mathbf{W}|D) = \frac{P(D|\mathbf{W}) \cdot P(\mathbf{W})}{P(D)} \]

where:
\begin{enumerate}
    \item \( P(\mathbf{W}|D) \) is the posterior distribution of the weights given the data \( D \).
    \item \( P(\mathbf{W}) \) is the prior distribution of the weights.
    \item \( P(D|\mathbf{W}) \) is the likelihood of the data given the weights.
    \item \( P(D) \) is the evidence or marginal likelihood of the data, often computed as an integral over all possible weights, which is usually intractable.
\end{enumerate}

Prediction in a BNN involves integrating over all possible weights, weighted by their posterior probabilities:

\[ p(y|\mathbf{x}, D) = \int p(y|\mathbf{x}, \mathbf{W}) \cdot P(\mathbf{W}|D) d\mathbf{W} \]

This integral is generally intractable due to the high dimensionality of the weight space, so approximate methods such as Markov Chain Monte Carlo (MCMC) or variational inference are used. The true posterior distribution \( P(\mathbf{W}|D) \) is approximated by a more tractable distribution \( q(\mathbf{W}|\theta) \), often chosen to be in a family of distributions that are easier to sample from, such as Gaussian distributions. The parameters \( \theta \) of this distribution are optimized using variational inference, which aims to minimize the divergence between \( q(\mathbf{W}|\theta) \) and \( P(\mathbf{W}|D) \).

\[ \text{KL}[q(\mathbf{W}|\theta) || P(\mathbf{W}|D)] \]
This is equivalent to maximizing the evidence lower bound (ELBO), which is often more computationally feasible than computing the posterior directly.
The objective function in variational inference is the Evidence Lower Bound (ELBO), which is given by:

\[ \text{ELBO}(\theta) = \mathbb{E}_{q(\mathbf{W}|\theta)}[\log P(D|\mathbf{W})] - \text{KL}[q(\mathbf{W}|\theta) || P(\mathbf{W})] \]

The first term is the expected log likelihood of the data under the approximate posterior, and the second term is the KL divergence between the approximate posterior and the prior distribution of the weights. To optimize the ELBO with respect to the variational parameters \( \theta \), we need to compute its gradient:

\[ \nabla_\theta \text{ELBO}(\theta) = \nabla_\theta \mathbb{E}_{q(\mathbf{W}|\theta)}[\log P(D|\mathbf{W})] - \nabla_\theta \text{KL}[q(\mathbf{W}|\theta) || P(\mathbf{W})] \]

To apply backpropagation, gradients of random variables (weights) with respect to the parameters \( \theta \) need to be computed. This is facilitated by the reparameterization trick, which expresses the random variables as deterministic functions of the parameters and some noise \( \epsilon \) that is independent of the parameters:

\[ \mathbf{W} = g(\epsilon, \theta) \]

where \( \epsilon \) is typically sampled from a simple distribution like a standard Gaussian, and \( g \) is a differentiable transformation that maps \( \epsilon \) and \( \theta \) to the space of weights. With reparameterization, the gradient of the expected log likelihood can be approximated by sampling and can be backpropagated through:

\[ \nabla_\theta \mathbb{E}_{q(\mathbf{W}|\theta)}[\log P(D|\mathbf{W})] \approx \frac{1}{L} \sum_{l=1}^{L} \nabla_\theta \log P(D|g(\epsilon^{(l)}, \theta)) \]

where \( L \) is the number of samples used in the Monte Carlo estimation.
Using gradient-based optimization methods like stochastic gradient descent (SGD) or Adam, the variational parameters \( \theta \) are updated iteratively:

\[ \theta_{\text{new}} = \theta_{\text{old}} + \alpha \nabla_\theta \text{ELBO}(\theta_{\text{old}}) \]

where \( \alpha \) is the learning rate.

Through this process, backpropagation in variational inference enables the optimization of the variational parameters \( \theta \) such that the approximate posterior \( q(\mathbf{W}|\theta) \) gets closer to the true posterior distribution \( P(\mathbf{W}|D) \), all while allowing the use of non-conjugate priors and complex models such as neural networks. This method is also known as Stochastic Variational Bayes because of the stochastic nature of the gradient estimation.

For inference, weights are sampled from an approximation to the posterior, and then predictions are made based on these samples rather than by performing the full integration over the weight space.
Once we have the approximate distribution \( q(\mathbf{W}|\theta) \), we can sample sets of weights \( \mathbf{W}^{(i)} \) from it:

\[ \mathbf{W}^{(i)} \sim q(\mathbf{W}|\theta) \]

These sampled weights \( \mathbf{W}^{(i)} \) are not the exact weights from the true posterior, but they are drawn from the distribution that we believe closely approximates the posterior.
For each set of sampled weights \( \mathbf{W}^{(i)} \), the BNN can make a prediction for an input \( \mathbf{x} \). This generates a distribution of outputs rather than a single point estimate:

\[ y^{(i)} = f(\mathbf{x}, \mathbf{W}^{(i)}) \]

where \( f \) is the neural network function. The full Bayesian prediction would integrate over the entire weight space, which is generally intractable. However, by sampling from the approximate posterior \( q(\mathbf{W}|\theta) \), we can estimate this integral by averaging the predictions from the sampled weights:

\[ p(y|\mathbf{x}, D) \approx \frac{1}{N} \sum_{i=1}^{N} p(y|\mathbf{x}, \mathbf{W}^{(i)}) \]

where \( N \) is the number of weight samples.

This is a Monte Carlo approximation of the integral, which is much more computationally feasible than the exact integration. The law of large numbers guarantees that, as \( N \) grows, the Monte Carlo approximation converges to the true integral.
The variance in the predictions from the different samples of weights provides a measure of uncertainty. If the variance is high, it means the model is less certain about its predictions.
This approach allows BNNs to make probabilistic predictions and quantify uncertainty without explicitly solving the intractable integral of the full Bayesian treatment. It is particularly useful in cases where the uncertainty estimate is as critical as the prediction itself, such as autonomous control systems in self driving cars.

\begin{figure}[tbhp]
\centering
   \includegraphics[width=0.6
   \linewidth]{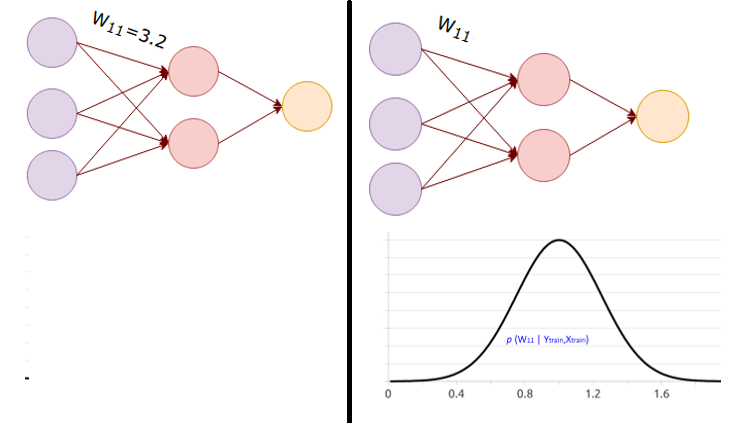}
   \caption{On the left the classical deep neural network and on the right is the Bayeisan Neural network ( As shown one of the parameter $w_{11}$ in DNN is a point value, however for the Bayesian Neural network, it is a Gaussian probability distribution}
   \label{fig:nn_train}
\end{figure}

\textbf{Deep Neural Network model, parameters and training}
DNN has wide ranging application from image recognition\cite{krizhevsky2017imagenet}, engineering design\cite{vardhan2021machine, kumar2023malaria ,vardhan2022deepal}, control design, anomaly detection\cite{vardhan2022reduced,sundaram2018predicting},etc. The basic architecture of DNN in this case is the one used in \cite{vardhan2022deep}. A DNN is based on a layered network of smaller computational units called artificial neurons.
Artificial neurons, which are tiny computing units, provide the basis of a multilayer network that powers a DNN. These neurons are fully linked between the input and output layers and are arranged in many layers. This experiment uses a fully connected feed-forward neural network, whose architecture is specified by $a=(L, N,\delta)$, where $L$ is the number of layers in the network. $L \in N$, where $l \in [L-1]$, $\delta$ is the activation function $\delta: R \mapsto R$, and $N$ is the set that reflects the number of neurons in each layer represented by $N_l$. A fully connected feed-forward six-layer neural network with $1$ and $\{256,128,64,32,16\}$ neurons in hidden layers was employed for this experiment. All hidden neurons have ReLU activation units and the output neuron is the linear activation unit.  The loss function, in this case, is the mean square error and the weight initialization is Gaussian normal distribution. The training is done for $500$ epochs by splitting the training and validation data into $90:10$ and early stopping is used to stop over-fitting.  

\begin{figure}[tbhp]
\centering
   \includegraphics[width=0.9
   \linewidth]{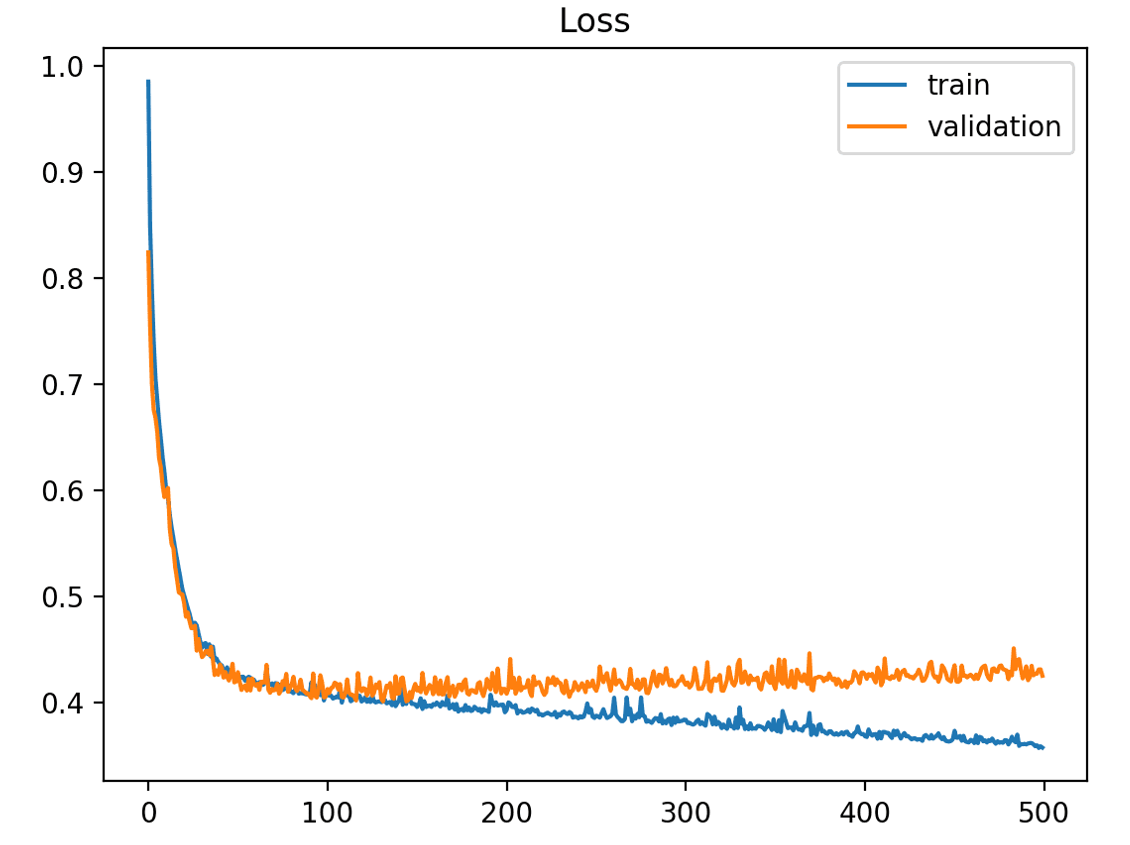}
   \caption{Training and validation loss for Bayesian neural network training}
   \label{fig:nn_train}
\end{figure}

\section{Experimental Results}
\label{sec:experimentalResults}
\begin{figure}[tbhp]
\centering
   \includegraphics[width=0.8
   \linewidth]{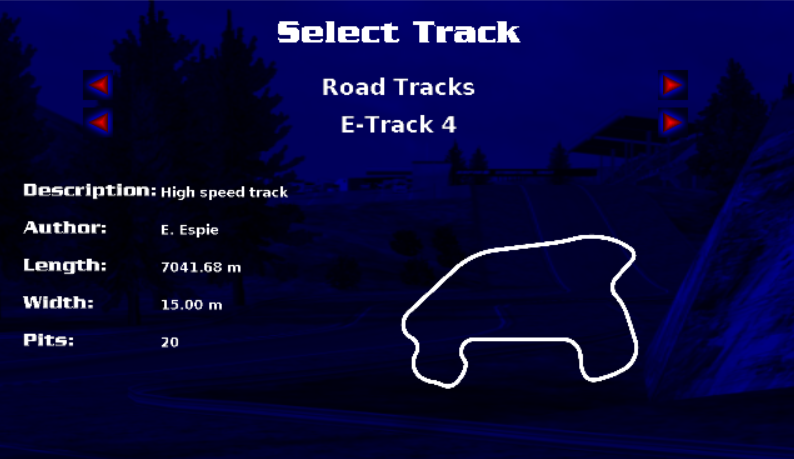}
   \caption{track: E-track 4}
   \label{fig:et4}
\end{figure}

Once the BNN models are trained, these models are tested on a similar but another track called 'E-Track 4' (refer to figure \ref{fig:et4}). The selection of the track is based on twists and turns and complexity in the track profile.  We set the target speed of the car to 60 miles/hrs. For the first experiment, we used both controllers to complete the track without any manual interventions.  The task was to finish one complete lap of this track. One complete lap of this track is 7.041 Km long with a track width of 15 meters.  The BNN controller was able to complete the task without any crash or off-the-track navigation.  

In the second experiment, we took a more difficult and track with more hairpin turns and twists. We monitored the variance in the steering value prediction and intervened and manually steered when the coefficient of variance becomes higher than the threshold.  
The coefficient of variance (CV) measure of the relative variability of a data set in relation to its mean and is calculated using following formula : 

\[ CV = \frac{\sigma}{\mu} \times 100 \]

where \(\sigma\) is the standard deviation of the data set and \(\mu\) is the mean of the data set.

The coefficient of variation is useful because it provides a dimensionless number that allows for comparison of the variability of data sets with different units of measurement or vastly different means. A lower CV indicates less variability (more consistency), while a higher CV indicates higher variability (less consistency). 

\begin{figure*}[h!]
\centering
   \includegraphics[width=1.0
   \linewidth]{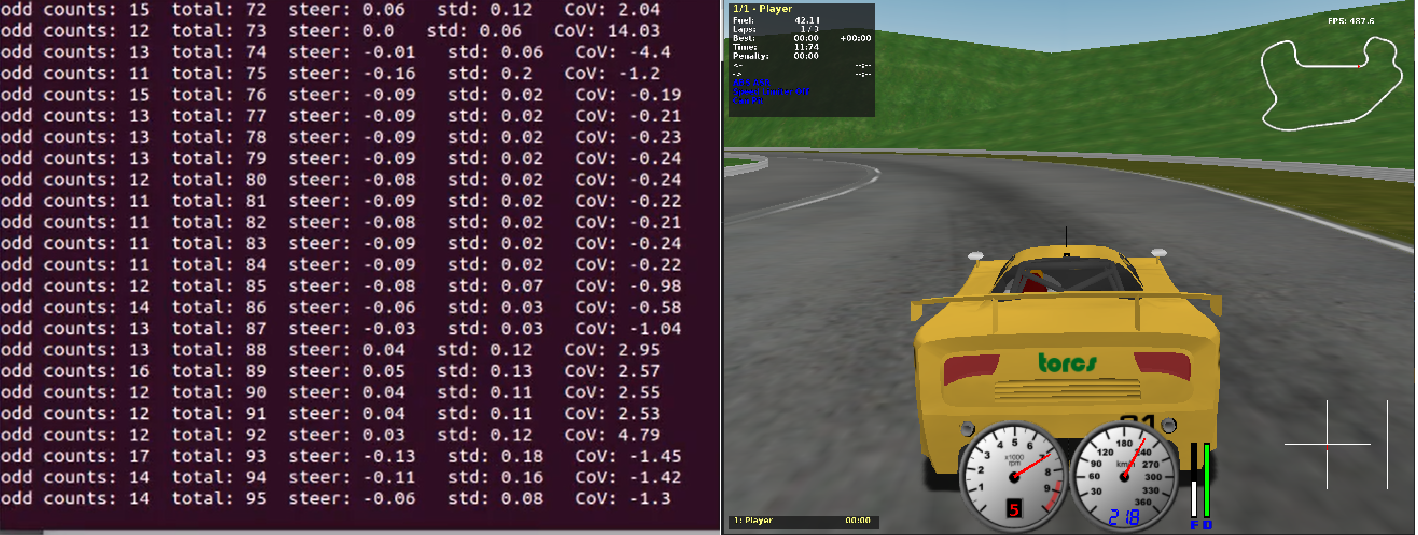}
   \caption{BNN controller in Torcs and related statistics, the statistics of our interest are: \textit{odd counts} -number of samples predictions out of 2 standard deviations from the mean, \textit{total}- total number of odd counts during whole simulation, \textit{steer}- steering value predicted by regressor, \textit{std}- standard deviation of prediction, \textit{CoV}- coefficient of variance, that explain the spread of the distribution. }
   \label{fig:drive}
\end{figure*}

 We observed that the trained random forest controller produced high CoV in scenarios which was very sharp turns and twist section of road that was never encountered on the training track. A snapshot of the BNN controller driving the car on the test track is shown in figure \ref{fig:drive}. 

\begin{figure}[tbhp]
\centering
   \includegraphics[width=0.8
   \linewidth]{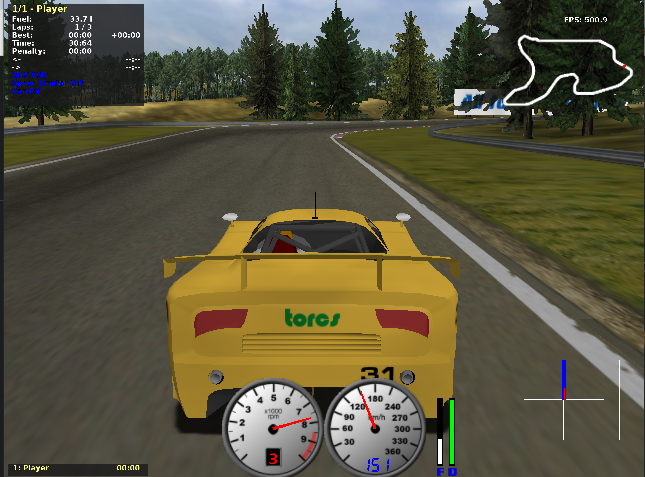}
   \caption{BNN controller being tested in Torcs track -Etrack6}
   \label{fig:et6}
\end{figure}
\begin{figure}[tbhp]
\centering
   \includegraphics[width=0.8
   \linewidth]{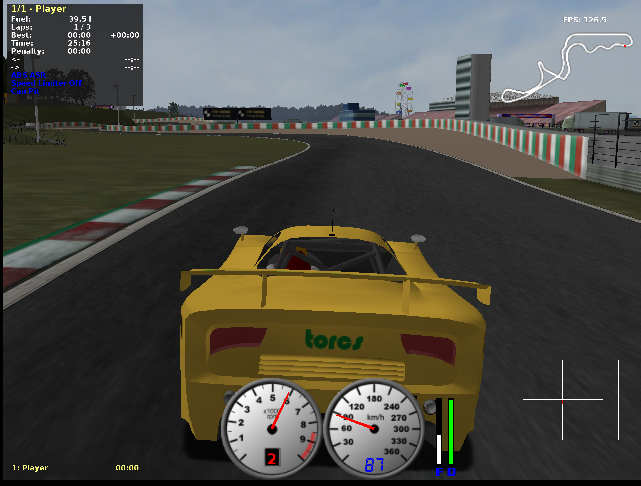}
   \caption{BNN controller tested on the track wheel2. It has some twisted turns as can be seen in the map on right -top corner. At only one hairpin turns, when controller uncertainty went beyond threshold (set to be the Coefficient of variance beyond $\pm 3$ continuously for 50 consecutive simulation steps (each step time is 2 milli seconds) then the manual control is kicked in. The manual control is relinquished again when the CoV comes within $\pm 3$ range for 50 consecutive steps. }
   \label{fig:wh2}
\end{figure}
\begin{figure}[tbhp]
\centering
   \includegraphics[width=0.8
   \linewidth]{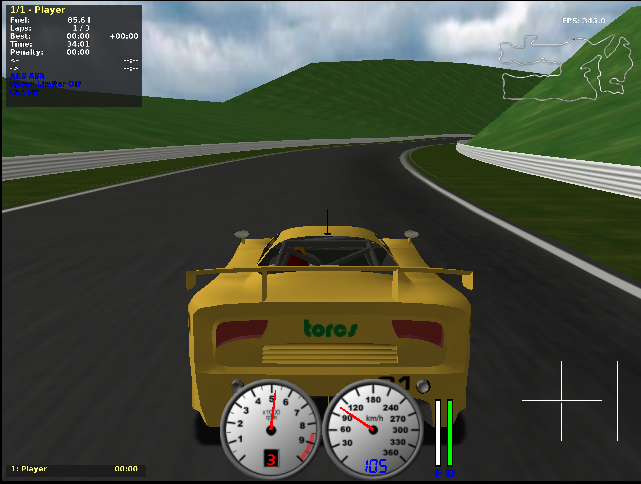}
   \caption{BNN controller tested on the track summer. It has highly twisted turns as can be seen in the map on right -top corner. At some hairpin turns, when controller uncertainty went beyond threshold (set the Coefficient of variance beyond $\pm 3$ continuously for 50 consecutive steps (each step time is 2 milli seconds) then the manual control is kicked in. The manual control is relinquished again when the continuous CoV comes within $\pm 3$ range for 50 consecutive steps. }
   \label{fig:summer}
\end{figure}
\section{Related Work}
\label{sec:relatedWorks}
The dominant approach to solving engineering problems are model-driven, like model-based design\cite{neema2019web}, model-based control\cite{brosilow2002techniques}, and model-based optimization\cite{vardhan2019modeling,neema2019design} etc. With the discovery of data-driven models especially deep neural networks the domain of model-based engineering is revolutionizing.  In recent times these data-driven models have been used and have worked well in image classification (AlexNet\cite{krizhevsky2017imagenet} ), engineering design\cite{vardhan2022data,moosavi2020role,vardhan2023search,volk2020biosystems}, autonomous driving car\cite{bojarski2016end,badue2021self,ni2020survey,shalev2017formal}, radiology \cite{gross1990neural}, human genome\cite{sundaram2018predicting}, and many more for developing state of the art control\cite{vardhan2022reduced,wu2019machine,tahsien2020machine} and prediction systems\cite{vardhan2022deep}.

According to research\cite{momtaz2018rate}, self-driving cars could potentially save up to $35,000$ lives annually in the United States alone by reducing traffic deaths by up to $99$ percent.
$93$ percent of car accidents are the result of driver mistake, according to information from 2007 US report\cite{facts2007us}, and rookie drivers—who seldom have more than $30$ hours of driving experience before receiving a license—are over-represented in fatal car accidents.
Moreover, crashes among senior drivers rise at the same time because they have a diminished capacity to assess their surroundings for unforeseen dangers, according to \cite{momtaz2018rate}.
Self-driving cars will have been trained using machine learning data-sets made up of a wide range of driving circumstances and driver behaviors for the equivalent of hundreds of hours behind the wheel before they hit the roads\cite{bojarski2016end}. The practical utility of AI-ML model is enormous \cite{vardhan2023fusion,dreossi2019verifai,vardhan2023constrained,vaishya2020artificial} and a better understanding and insight about the model's performance would be useful for practical control designer. Since Deep learning is a data hungry model, it is not possible to generate an enormous amount of data for each application. In such cases, this DNN models fail to perform and alternative learning model can be useful.

\section{Conclusion and Future Work}
\label{sec:conclusionFutureWork}

\subsection{Conclusion}
 In this work, we designed a controller for the lateral control of a vehicle using AI-ML models. Experiments' results show that:
 \begin{enumerate}
     \item  Bayesian neural network based controller is trained and tested in different track and environment using Torcs simulator. Due to probabilistic nature of network, BNN can predict uncertainty in prediction and can be used to make decision about the reliability or confidence of the control action.
     \item Trained on one track and deployed on other tracks, the BNN-based controller was able to complete some of the the track without crashing and with out manual interventions. In some highly twisted tracks , sometime manual intervention was invoked based on threshold set and prediction confidence interval. 
 \end{enumerate}

\subsection{Future Work}
Our experiments are use cases to understand deeper about possibility of using probabilistic AI model control of safety critical system. For future work, we want to compare the earlier work \cite{fogla2023vehicle} with current one based on various performance metrics.

\bibliographystyle{IEEEtran}
\bibliography{references}
\end{document}